\documentclass{article}

    \usepackage[preprint]{neurips_2021}

\usepackage[utf8]{inputenc} 
\usepackage[T1]{fontenc}    
\usepackage{hyperref}       
\usepackage{url}            
\usepackage{booktabs}       
\usepackage{amsfonts}       
\usepackage{nicefrac}       
\usepackage{microtype}      
\usepackage{xcolor}         
\usepackage{epsfig}
\usepackage{graphicx}
\usepackage{amsmath}
\usepackage{amssymb}
\usepackage{multirow}

\DeclareMathOperator*{\argmax}{arg\,max}

\def\etal{\emph{et al}.}

\def\b{\textbf{b}}

\def\1{\textbf{1}}

\def\i{\textbf{i}}
\def\n{\textbf{n}}
\def\b{\textbf{b}}

\def\tt{\texttt{}}

\title{Visual Transformer for Task-aware Active Learning}

\author{Razvan Caramalau$^1$, Binod Bhattarai$^1$ and Tae-Kyun Kim$^{1,2}$ \\
$^1$Imperial College London, UK\\
$^2$KAIST, South Korea\\
{\tt\small \{ r.caramalau18, b.bhattarai, tk.kim\}@imperial.ac.uk}
}

\begin{document}

\maketitle

\begin{abstract}
 Pool-based sampling in active learning (AL) represents a key framework for annotating informative data when dealing with deep learning models. In this paper, we present a novel pipeline for pool-based Active Learning. Unlike most previous works, our method exploits accessible unlabelled examples during training to estimate their co-relation with the labelled examples. Another contribution of this paper is to adapt Visual Transformer as a sampler in the AL pipeline. Visual Transformer models non-local visual concept dependency between labelled and unlabelled examples, which is crucial to identifying the influencing unlabelled examples. Also, compared to existing methods where the learner and the sampler are trained in a multi-stage manner, we propose to train them in a task-aware jointly manner which enables transforming the latent space into two separate tasks: one that classifies the labelled examples; the other that distinguishes the labelling direction. We evaluated our work on four different challenging benchmarks of classification and detection tasks viz. CIFAR10, CIFAR100, FashionMNIST, RaFD, and Pascal VOC 2007.  Our extensive empirical and qualitative evaluations demonstrate the superiority of our method compared to the existing methods. Code available: \url{https://github.com/razvancaramalau/Visual-Transformer-for-Task-aware-Active-Learning} 
\end{abstract}

\section{Introduction}
%
%
In the recent success stories of Deep Learning in image classification~\cite{krizhevsky2012imagenet, he2016deep, vi} and object detection ~\cite{ssd, objdet2018sota, objdet2020sota} the large-scale  labelled data sets have been crucial. 
Data annotation is a time-consuming task, needs experts and is also expensive. Active Learning~\cite{near2,Pinsler2019BayesianApproximation,pu2016variational,ugcn,caramalau2021active} 
is getting popular to select a subset of discriminative examples incrementally to learn the model for 
downstream tasks. In the AL frameworks usually, a learner,  a sampler, and an annotator complete a loop and repeat the cycle. In brief, a learner minimizes the 
objective of the downstream task and the sampler selects the representative unlabelled examples given a fixed annotation budget. And, an annotator queries the labels of 
the unlabelled data recommended by the sampler. Based on the category of the employed sampler, the whole paradigm of AL can be broadly dissected into
uncertainty based~\cite{Houlsby2011BayesianLearning, Gal2017DeepDatab, Kirsch2019BatchBALD:Learning}, geometric based~\cite{wolf,Sener2017ActiveApproach}, model based~\cite{Yoo2019LearningLearning, Sinha2019VariationalLearning, cdal, csal}, and so on.



In this paper, we focus on model based~\cite{Sinha2019VariationalLearning,ugcn,Yoo2019LearningLearning} active learning pipelines for pool-based sampling. 
The importance of this type of pipeline is growing and 
getting more relevant than ever before due to the increasing use of deep learning algorithms. In this scenario, given a large volume of unlabelled data, an initial model is trained on a small subset of randomly annotated examples.
In the later stages, the samples are annotated in the guidance of the model trained in the previous stage. The fate of this method category is determined by 
the performance of initial model which is commonly known as \emph{cold-start} problem~\cite{csal}. To tackle such a problem and improve the performance 
of the model in its early stage, training the model in a semi-supervised fashion is slowly getting into attention~\cite{csal}. Exploiting unlabelled data
along with the labelled data in joint or a multi-task setup improves the generalization of the model~\cite{caruana1997multitask}. However, the previous work
minimizes the loss functions such as consistency loss~\cite{csal} which is indirect to the downstream task i.e. class category loss.
Also, these methods rely only on exploiting unlabelled data to improve the performance of the model. Unlike these methods, our approach tackles this
problem from both aspects i.e. making use of unlabelled data as well as engineering the model architecture. To this end, 
we adapted Visual Transformer~\cite{vt} in the pipeline with the learner and exploited the unlabelled data for better generalisation. Finally, we train the model in a joint-learning framework by minimizing a labelled vs unlabelled discriminator (sampler) as well as the downstream task-aware loss.

Visual Transformers (VT)~\cite{vt} are attaining state-of-the-art results on various tasks such as image classification~\cite{vi, vt}, detection~\cite{objdet2020tf}, and so on. 
To the best of our knowledge, this is the first work adapting VT for active learning framework.  Figure~\ref{fig:pipeline} demonstrates the pipeline of 
the proposed method. A batch of both the unlabelled and labelled examples are passed through few convolutional layers sequentially. The output batch from
these layers are fed into the VT layers at once. CNN layers are myopic in nature by extracting the statistics of local information as image features.
Uncertainty on such feature space helps us to select images with variations on blurriness, contrasts, textures etc. However, a non-local interaction between the 
unlabelled and labelled examples is essential to identify the complimentary examples to query their labels. To this end, we propose to integrate VT in between CNN layers and the output layer. Previous works on VT for computer vision~\cite{vi} divided the images by a regular grid to extract local patches and fed them to VT
to extract the non-local interactions between parts of an image. As our task is to identify the most discriminative images, hence, we consider each level representation of both labelled and unlabelled examples as an input channel to this module. Thus, VT extracts non-local interactions between the labelled-unlabelled examples while uncertainty on
such feature space allows us to select the images which are sufficiently different on  a visual concept. The output of the VT is feed to output which is bifurcated into labelled vs unlabelled discriminator and task-specific auxiliary classifier. Uncertainties on feature space to minimizing these two losses help us
to find the unlabelled examples which are sufficiently different to labelled examples and relevant to the downstream task. This address the problem of earlier
methods~\cite{Sener2017ActiveApproach} selecting examples from high-density regions irrespective of the decision boundary.

We summarise the contributions of this paper in the following bullet points:

\begin{itemize}
    \item We propose a novel task-aware joint-learning framework for active learning.
    \item We adapted Visual Transformer for the first time in the pipeline of active learning.
    \item We evaluated our methods for sub-sampling real and synthetic examples for four different image classification and one object detection benchmarks.
    \item We outperformed existing methods by a large margin and attain a new state-of-the-art performance. 
\end{itemize}

\section{Related Works}
\label{related_works}

The current taxonomy of active learning is founded on the extensive survey of Settles \cite{settles.tr09}. This gathers all the classical approaches together with the three scenarios of active learning. Most deep learning works, including ours, rely on the pool-based scenario. Depending on the mechanisms used for sampling or deriving heuristics of the unlabelled data, we can categorise methods as uncertainty-aware, geometric representation and model-based. The first category has been initially applied with the Monte Carlo (MC) Dropout approximation for deep Bayesian models of Gal \etal \cite{Gal2016DropoutGhahramani}. Thus, the selection in the active learning study \cite{Gal2017DeepDatab} is inspired from classical approaches as maximum entropy \cite{entropy} or BALD \cite{Houlsby2011BayesianLearning}. Another approach of gathering the uncertainty from models is by querying a committee machine \cite{settles.tr09}. A recent work that expanded to deep learning this classic principle has been presented by Beluch \etal \cite{BeluchBcai2018TheClassification}. This method overcame in performance the works centred on the MC Dropout mechanism. However, with the complexity increase of current deep learning models, both iterative approaches have proven to be hardly applicable.

On this concern, the second category tackles this issue by evaluating geometrically the representations of the downstream task. We acknowledge Senner and Savarese \cite{Sener2017ActiveApproach} as the most representative work with the CoreSet algorithm. Their methodology evaluates a global fixed radius to cover the feature space by selecting a subset of unlabelled. This has been succeeded by other learners of \cite{near2,coresvm,Har-Peled2007}. We include this competitive baseline for comparison in our experiment section.

The third category, and the most recent one, deploys dedicated learning models, also referred to as samplers, for querying new data. The first proposed module, Learning Loss \cite{Yoo2019LearningLearning}, tracks uncertainty by training end-to-end and estimating the predictive loss of the unlabelled. The modular aspect of this category permits deploying the samplers to diverse applications. Our method inherits this advantage. Sinha \etal \cite{Sinha2019VariationalLearning} defined the sampler training framework VAAL separately where a variational auto-encoder (VAE) maps all the available data in a latent space. The selection principle is based on the adversarially trained discriminator between labelled and unlabelled. The fallback of this method, lack of task-awareness, has been addressed in \cite{Zhang2020State-RelabelingLearning, cdal,caramalau2021active}. Hence, Agarwal \etal \cite{cdal} proposes CDAL that combines the sampler with the contextual diversity while enlarging the receptive feature domain. Following similar trends, Caramalau \etal \cite{caramalau2021active} deploys Graph Convolutional Networks (GCNs) for feature propagation between labelled and unlabelled images.  These works are close to ours and are evaluated in the experiment section.

Because our proposed AL framework combines the semi-supervised learning (SSL) strategy with a visual transformer, we further investigate related literature. SSL and AL have been recently considered for deep learning in a few works \cite{csal, Sener2017ActiveApproach, sslalspeech, curmat}. The first attempt is noted in CoreSet \cite{Sener2017ActiveApproach} during AL cycles. A more elaborated method CSAL analyses the consistency loss of unlabelled data when trained end-to-end. Thus, it achieves state-of-the-art on the image classification datasets. Furthermore, we outperform this baseline in our analysis under their experiment settings. On the other end, the visual transformer has been initially designed for vision applications by Dosovitskiy \etal \cite{vi}. They considered the natural language processing BERT \cite{bert} approach on patches of the images to learn the non-local representations. Following this work, transformers have successfully replaced convolutional layers in \cite{vt} while boosting the accuracy under a similar number of parameters. Our methodology is supported by the insights from this research.
\section{Method}
\begin{figure*}
    \centering
    \includegraphics[trim=0cm 0cm 0cm 0cm, clip, width=1.0\textwidth]{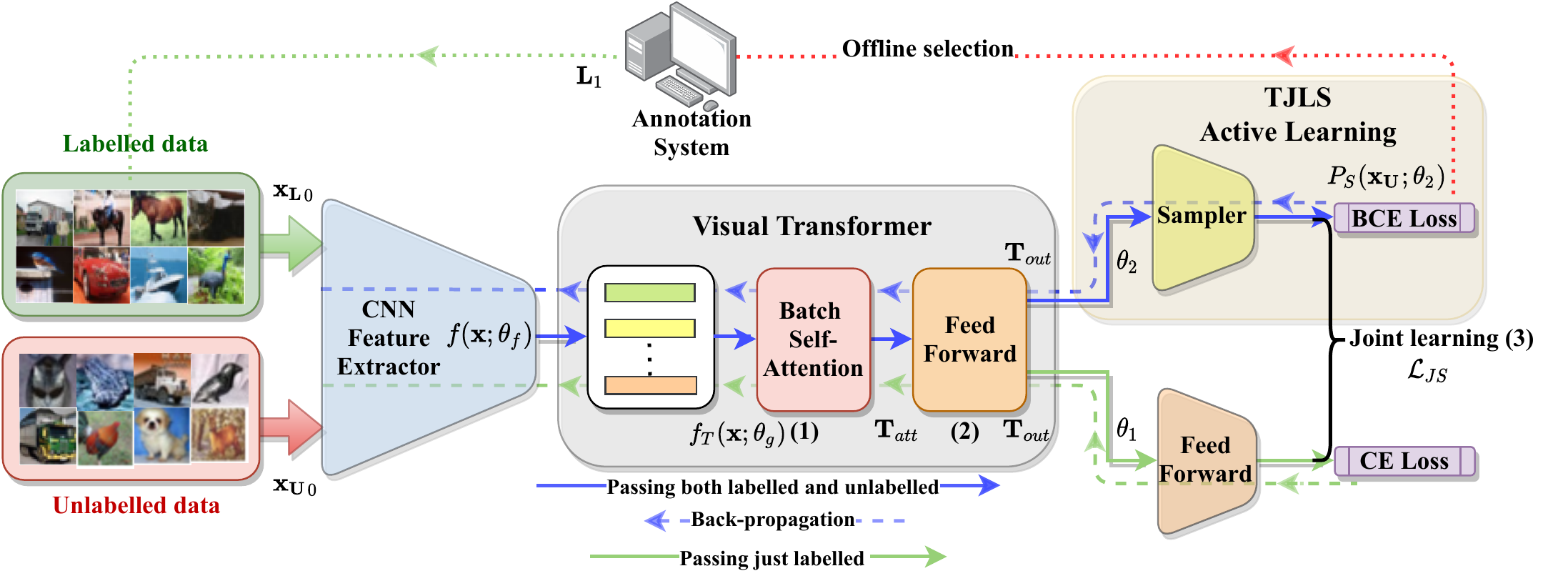}
    \caption{This diagram depicts the proposed pipeline. We pass both the labelled and unlabelled examples through the same CNN and extract the visual features of each image encoded within the batch.
    These representations are fed to the Visual Transformer. Its outputs are passed to the bifurcated branches to minimize both the learner's 
    (class cross-entropy ) and the sampler's objectives  (binary cross-entropy). }
    \label{fig:pipeline}
\end{figure*}


In this Section, we start with the formal definition of pool-based Active Learning in general followed by our contributions on introducing VT in the 
pipeline and the task-aware joint-learning objective.
Given a large pool of unlabelled data $\mathbf{x_U} \in \mathbf{U}_{pool}$, the pipeline begins with a cold-start training of the model by randomly selecting a 
small subset and labelling an initial set $\mathbf{x_L}_0, \mathbf{y_L}_0 \in \mathbf{L}_{0} \subset \mathbf{U}_{pool}$. The performance of the initial model
is crucial for the end outcome of the framework~\cite{csal} for model-based active learning. Our contribution lies in addressing this problem which is 
also commonly known as \emph{cold-start} problem. To this end, we took two approaches: jointly learning the parameters of the learner and sampler utilising the accessible unlabelled examples and adapting the Visual Transformer as a bottleneck of the pipeline.
If $\b$ is the budget to sample unlabelled examples over multiple selection stages, 
the main objective of the pool-based AL scenario is to obtain fast generalisation of the learner with the least number of labelled 
subsets $\n$ in order to keep minimum samples. $\forall \i \in \n $,  $\mathbf{L}_i = (\mathbf{x_L}_i, \mathbf{y_L}_i) \subset \mathbf{U}_{pool}$  
represents the annotated examples in every $\i^{th}$ subset.





Figure~\ref{fig:pipeline} outlines the proposed pipeline. From the Figure, we can see that both the labelled and unlabelled examples are fed to the 
image feature extractor, $f$.  For most vision applications, the backbone of the learner is a feature extractor commonly formed of CNNs such as ResNet~\cite{he2016deep}
and VGG architectures~\cite{Simonyan2015VeryRecognition}. From the initial labelled set, $\mathbf{x_L}_0, \mathbf{y_L}_0 \in \mathbf{L}_{0}$ and   $\mathbf{U}_{0} \in \mathbf{U}_{pool} \setminus \mathbf{L}_{0}$, unlabelled examples,
we infer them to the feature extractors. In our case, we take an equal number of labelled and unlabelled examples to balance the number of training examples.
In a batch of $B$ examples, we choose both unlabelled and labelled examples and feed them into $f(\theta_f)$
where $f:\mathbb{R}^{B \times WH\times C}  \to \mathbb{R}^{B \times wh \times |F|}$. Here, $W,H$ are height and width of the images, $C$ is channels of input images,
$w,h$ are width and height of channels from the last convolutional layer, $|F|$ is the total number of filters on the last convolutional layer of feature extractor.
Earlier methods~\cite{ugcn,Yoo2019LearningLearning} feed the output of the convolutional layers to the output
layers to minimize the loss. Instead, we feed the output of these layers to the Visual Transformer before feeding to the output layers. 
CNN layers handle only local dependencies, but non-local dependencies between the labelled and unlabelled examples are essential to select 
the complementary unlabelled examples. 
Visual Transformer has been quite successful in modelling non-local dependencies. 
UncertainGCN~\cite{ugcn} uses GCN to handle long-range dependencies in Active Learning. A comparative study on GCN and self attention~\cite{vt}  has been 
shown that the self-attention aggregation function retains better the diverse concepts than the GCN.

\noindent \textbf{Visual Transformer.}
\cite{vi} is the first work to apply Visual Transformer successfully for image classification.  In this work, each image is divided by regular grids into patches and 
each of the patches is considered input tokens to the transformer. Another following work~\cite{vt} compressed the features of CNN backbones to visual tokens while 
the transformer acted as the final convolutional layer. Inspired by these works, we plugged a visual transformer as a neck between the feature extractor and 
the output layer. In our case, the inputs to the transformer are batches of feature maps that we obtained from the feature extractors 
as $f:\mathbb{R}^{B \times WH\times C}  \to \mathbb{R}^{B \times wh \times |F|}$.

Different from \cite{vi, vt}, we foresee the AL framework to be least intrusive to the learner's architecture. This allows the methodology to be plugged into various designs and applications. We, therefore, do not further post-process the output of CNN feature extractor. Similarly to GCNs \cite{kipf2016semi}, we want to explore the relationships between the nodes of the graph. However, given our input $f(\mathbf{x}; \theta_f)$ ($\theta_f$ are the parameters of the feature extractor), we propose to deploy the transformer blocks within the batch $B$. Consequently, all the channels of the feature maps from each batch are considered as tokens to VT. Although in the standard architecture \cite{attall}, the inputs to the transformer are positional encoded, this does not apply in our scenario where the order of the images is irrelevant. As mentioned in the Introduction, our objective is to extract non-local relationships between the images not within the image.  To simplify furthermore the transformer's architecture, we exclude the decoder part as the target domain is absent in our case.

From the seminal work on self-attention~\cite{attall}, the main building blocks of the transformer's encoder are a batch self-attention block and a point-wise
feed-forward network with residuals and layer normalisation. For the self-attention, we transpose to the batch and concatenate the feature maps so that $f_T(\mathbf{x}; \theta_f) \in \mathbb{R}^{wh|F| \times B}$. The key, query and value matrices ($\mathbf{W}_{\mathbf{K}_{att}}, \mathbf{W}_{\mathbf{Q}_{att}}, \mathbf{W}_{\mathbf{V}_{att}}$) are packed together to model the interactions between the features. This also favours inner-domain relationships as batch normalisation is commonly used in CNNs for regularisation. The operations of the batch self-attention are summarised as follows:

\begin{equation}
    \mathbf{T}_{att} = f_T(\mathbf{x}; \theta_f) + \frac{\textrm{softmax}((f_T(\mathbf{x}; \theta_f)\mathbf{W}_{\mathbf{K}_{att}})(f_T(\mathbf{x}; \theta_f)\mathbf{W}_{\mathbf{Q}_{att}})^T)}{\sqrt{dim_{\mathbf{K}_{att}}}}(f_T(\mathbf{x}; \theta_f)\mathbf{W}_{\mathbf{V}_{att}}),
    \label{eq:2}
\end{equation}

where $\mathbf{T}_{att}$ and $\textrm{softmax}$ are the output of the batch self-attention layer and its activation function. While keeping the dynamics within the batch, we pass $\mathbf{T}_{att}$ through the point-wise feed-forward network. We define $\mathbf{T}_{out} (f_T; \theta_g)$ the output of this block, with $\theta_g$ as all VT parameters. The following equation underlines its processes:

\begin{equation}
    \mathbf{T}_{out} = \mathbf{T}_{att} + \sigma(\mathbf{T}_{att}\mathbf{W}_{h_1})\mathbf{W}_{h_2},
    \label{eq:3}
\end{equation}

where $\mathbf{W}_{h}$ are the weights from the feed-forward network and $\sigma$ represents the sigmoid activation function.

\noindent \textbf{Task-Aware Joint-Learning.}
Recently, the state-of-the-art for deep active learning has been gained by model-based methods like Learning Loss\cite{Yoo2019LearningLearning}, VAAL\cite{Sinha2019VariationalLearning}, CDAL\cite{cdal} and UncertainGCN\cite{ugcn}. These fundamentally comprise dedicated trainable models to sample unlabelled data. Apart from Learning Loss, the downfall of these methods is the sub-optimal multi-stage training processes. Also, in limited budget scenarios, the risk of over-fitting appears from the initial cold-start sampling. Unlike the approaches of these methods, we proposed to optimise the parameters jointly. Also, customize the objective depending upon the downstream tasks. In our case, we have considered image classification and object detection. But, our method can be 
easily extended to other tasks.

The representations of the labelled examples and unlabelled examples from the transformer
are fed into the bifurcated branch of the network. One of them minimizes the binary cross-entropy loss to distinguish labelled examples from unlabelled examples. Another branch minimizes the task-specific loss. For the classification task, we minimize class categorical loss. 
Similarly, for object detection, we minimize the combination of confidence and localisation loss as stated in~\cite{ssd}.
Thus the overall objective of the network when the downstream task is classification task is as shown in the Equation~\ref{eq:1}. In the same manner, the overall objective becomes as stated in Equation~\ref{eq:1_1} when the downstream task is object detection. In the Equations, 
$\lambda_1, \lambda_2$ are the weighting factors corresponding to each task. $\theta_1$ and $\theta_2$ are the learnable parameters of the main task and the sampler branch, respectively.




To learn them, we apply gradient back-propagation. We alternate the gradient between the sampler and the task branch for every batch of data as shown by the backward head arrow in Figure~\ref{fig:pipeline}. This adds an inductive bias while avoiding over-fitting the data
or the random noise. An elaborated deduction was presented by Goyal \etal in \cite{inductivebias}. 

    \begin{align}
    \mathcal{L}_{JL} = \lambda_1 \mathcal{L}_{CE}(\mathbf{x_L}_0, \mathbf{y_L}_0; [\theta_1, \theta_g, \theta_f] ) + \lambda_2 \mathcal{L}_{BCE}(\mathbf{x_L}_0, 1 ; [\theta_2, \theta_g, \theta_f]), 
    \label{eq:1_1}
    \\
    \mathcal{L}_{JL} = \lambda_1 \mathcal{L}_{SSD}(\mathbf{x_L}_0, \mathbf{y_L}_0; [\theta_1, \theta_g, \theta_f] ) + \lambda_2 \mathcal{L}_{BCE}(\mathbf{x_L}_0, 1 ; [\theta_2, \theta_g, \theta_f]),
    \label{eq:1}
    \end{align}





\noindent \textbf{Sampling the unlabelled data.}
To recap, the proposed AL framework trains jointly both learner and sampler. As a bottleneck between feature extractor and output layers, we add a visual transformer between the feature extractor and the two task branches. Combining the unlabelled and labelled data helps in preserving the most meaningful features during the learning stage. Moreover, the transformer will be exposed also to the unlabelled data dependencies. This happens as the task of the sampler is to classify the labelled from the unlabelled. The two are categorised as 1 or 0, respectively. If the unlabelled examples $\mathbf{x_U}_0 \in \mathbf{U}_{0}$ are easily differentiated by the sampler, we want to target for selection the most uncertain $\mathbf{x_U} \in \mathbf{U}_{pool}$.

Similarly, it has been done in an adversarial manner by VAAL \cite{Sinha2019VariationalLearning}, although their AL framework is not linked with the main task. Therefore, we derive our selection principle from UncertainGCN\cite{ugcn}. Given a budget of $\b$ points, we infer the entire unlabelled pool and select the samples with the lowest posterior probability of the discriminator branch. We notate $P_S$ as the confidence score of the posterior. Considering the first selection stage, we can evaluate the new labelled set $\mathbf{L}_1$ with:

\begin{equation} 
\label{eq:4}
    \mathbf{L}_1 = \mathbf{L}_0  \cup \argmax_{i=1\cdots b} P_S( \mathbf{x_U}; \theta_2).
\end{equation}

We compute the  $\argmax$ because the highest confidence score for the unlabelled is when $P_S( \mathbf{x_U}; \theta_2)$ is closer to 0. This selection process along with re-training is repeated until the targeted performance of the downstream task is reached.

For convenience, we denote the proposed AL framework as \emph{TJLS}( Transformer with Joint-Learning Sampler). In the next part, we thoroughly quantify the stated method and motivations. Furthermore, to observe the impact of the transformer bottleneck, we also investigate the pipeline \emph{JLS} without it.  

\section{Experiments}
Here we present both the quantitative (including ablation studies) and qualitative evaluations in a detailed manner. We employed our method
TJLS (visual transformer in the joint-learning sampler) for two different tasks: image classification and objection. 
\paragraph{Baselines.} 
We choose state-of-the arts from different categories like uncertainty-based (MC Dropout \cite{Gal2016DropoutGhahramani}, DBAL \cite{Gal2017DeepDatab}), geometric (CoreSet \cite{Sener2017ActiveApproach}) and the most recent model-based (Learning Loss \cite{Yoo2019LearningLearning}, VAAL\cite{Sinha2019VariationalLearning}, CDAL\cite{cdal}, CSAL\cite{csal}). 

The standard practice to acquire datasets is through random sampling from the uniform distribution. This does not require any active learning mechanism for selection. The first methods to explore the uncertainties in deep learning models have been MC Dropout and, its extension, DBAL. Both approximate the learner in a Bayesian fashion.
The two uncertainty-based methods differ by their selection criteria. MC Dropout relies on maximum entropy, while DBAL maximises information with BALD \cite{Houlsby2011BayesianLearning}. From a geometric perspective, the most successful work, CoreSet \cite{Sener2017ActiveApproach}, estimates risk minimisation between the labelled set and a core points of unlabelled. Fundamentally, a k-centre Greedy \cite{wolf} algorithm measures the distances between labelled and unlabelled learner's features. 

As our method falls in the third category, model-based, we assess four recent state-of-the-art: Learning Loss \cite{Yoo2019LearningLearning}, VAAL\cite{Sinha2019VariationalLearning}, CDAL\cite{cdal}, CSAL\cite{csal}. The first work to introduce learnable parameters for sampling is Learning Loss. Similarly, to our work, they train end-to-end the learner with the sampler, but the extra module predicts the downstream task loss. A more complex sampler has been proposed by VAAL where a variational auto-encoder is trained in an adversarial manner with both labelled and unlabelled data. The selection principle is close to ours by picking the hardly discriminated samples. However, the sampler is not task-aware with the main objective. On the other hand, CDAL provides a reinforcement learning module for a Bi-LSTM \cite{lstm} that captures the contextual diversity of the learner. Their selection criteria keeps the unlabelled batches with the highest reward. The last baseline that we tackle in our experiments is CSAL. This method also optimizes the task model with unlabelled samples as in JLS. The main difference to ours consists in using augmented unlabelled data to compute a consistency loss that does not overlap with the end task. 

\subsection{Image Classification}
\noindent \textbf{Datasets and Implementation Details.} For image classification, we took three well known data sets: CIFAR-10, CIFAR-100 \cite{cifar} and FashionMNIST\cite{Xiao2017Fashion-MNIST:Algorithms}. 
The training set of each of the dataset makes $\mathbf{U}_{pool}$ . CIFAR-10 and CIFAR-100 consists of RGB images whereas FashionMNIST 
is grey-scale.

From an architectural perspective, we swap the main deep CNN model between VGG-16 \cite{Simonyan2015VeryRecognition} and ResNet-18 \cite{he2016deep}. These learners 
are combined with the visual transformer and the sampler of our method TJDS. We set the hidden dimension of the transformer block for both self-attention and
feed-forward module to 128. The label discriminator part of the sampler is composed of two layers. The hidden units of both of them
are 512. We fixed these values for our pipeline for these experiments. Part of the hyper-parameter tuning behind this configuration is discussed 
later in this section. In Equation \ref{eq:1}, we weight the impact of the two losses by $\lambda_1$ and $\lambda_2$. From cross-validation, we notice 
that setting the weight of the sampler ($\lambda_1$) to 50\%  of the auxiliary loss ($\lambda_2$) brings more stability in joint-learning. 
However, $\lambda_1$ is set to 1 when labelled images are inferred through the downstream task.
During training, we fix the batch size to 128 for all the methods. We optimize the gradient of the joint-learning with SGD by setting a learning rate of 0.01,
a weight decay of 5e-4, and epochs to 200. To measure the performance of the AL framework, we evaluate the mean accuracy over 5 trials for 
7 selection stages. 
\\
\noindent\textbf{Quantitative analysis.} We deploy VGG-16 for the CIFAR-10 and CIFAR-100 experiments. We start with an initial labelled set $\mathbf{x_L}_0$ with 10\% of the original training set. For the selection phase, we allocate a budget $\b$ of 5\% every time identical to that of VAAL\cite{Sinha2019VariationalLearning} or CDAL\cite{cdal}. Our TJLS or JLS algorithm requires a subset $\mathbf{x_U}_0$ of unlabelled examples to train the sampler branch. In these benchmarks, we vary this set equal to the percentage of labelled. Following this, the sampler is not biased towards any group. 

\begin{figure*}
    \centering
    \includegraphics[trim=2cm 0cm 2cm 0cm, clip, width=.32\textwidth]{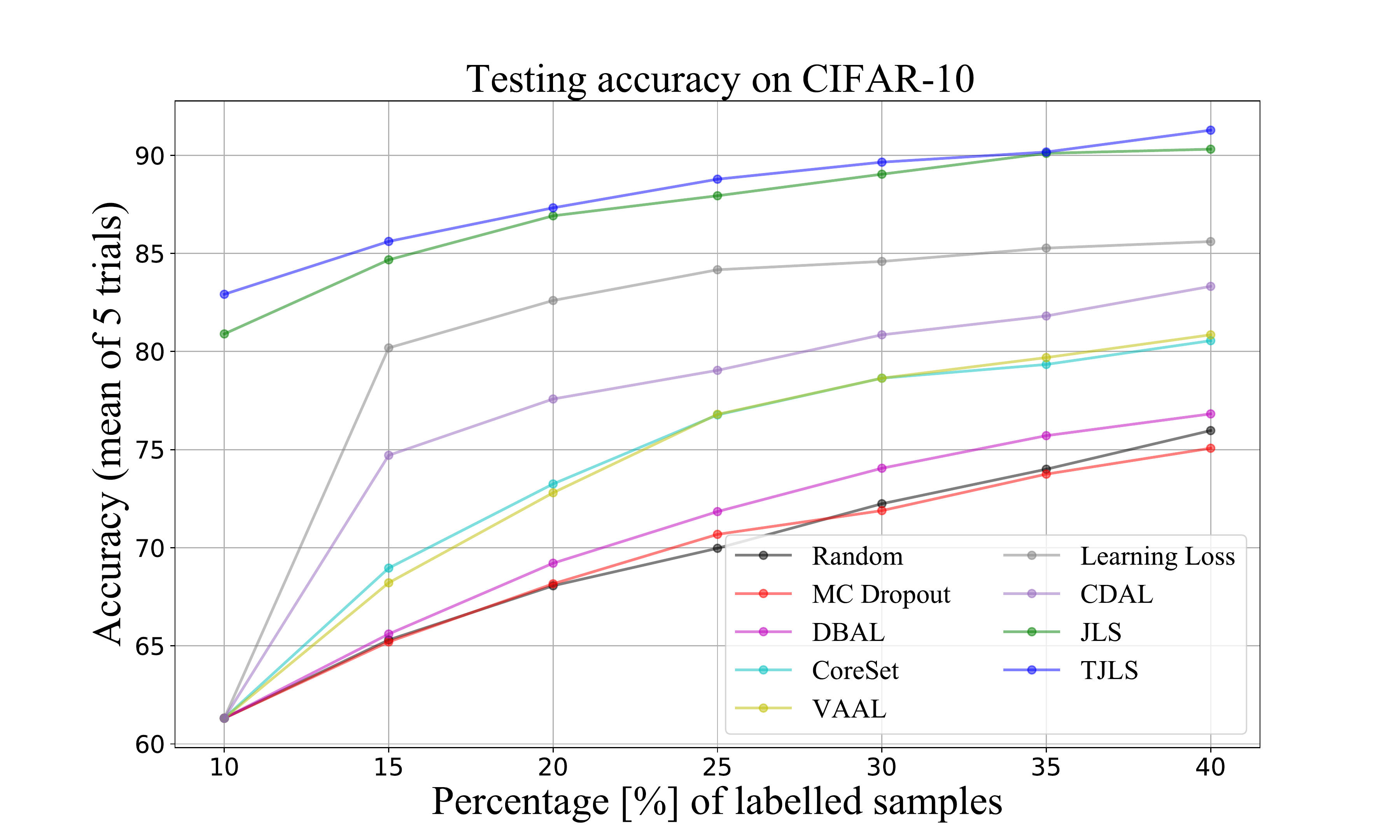}
    \includegraphics[trim=2cm 0cm 2cm 0cm, clip, width=.32\textwidth]{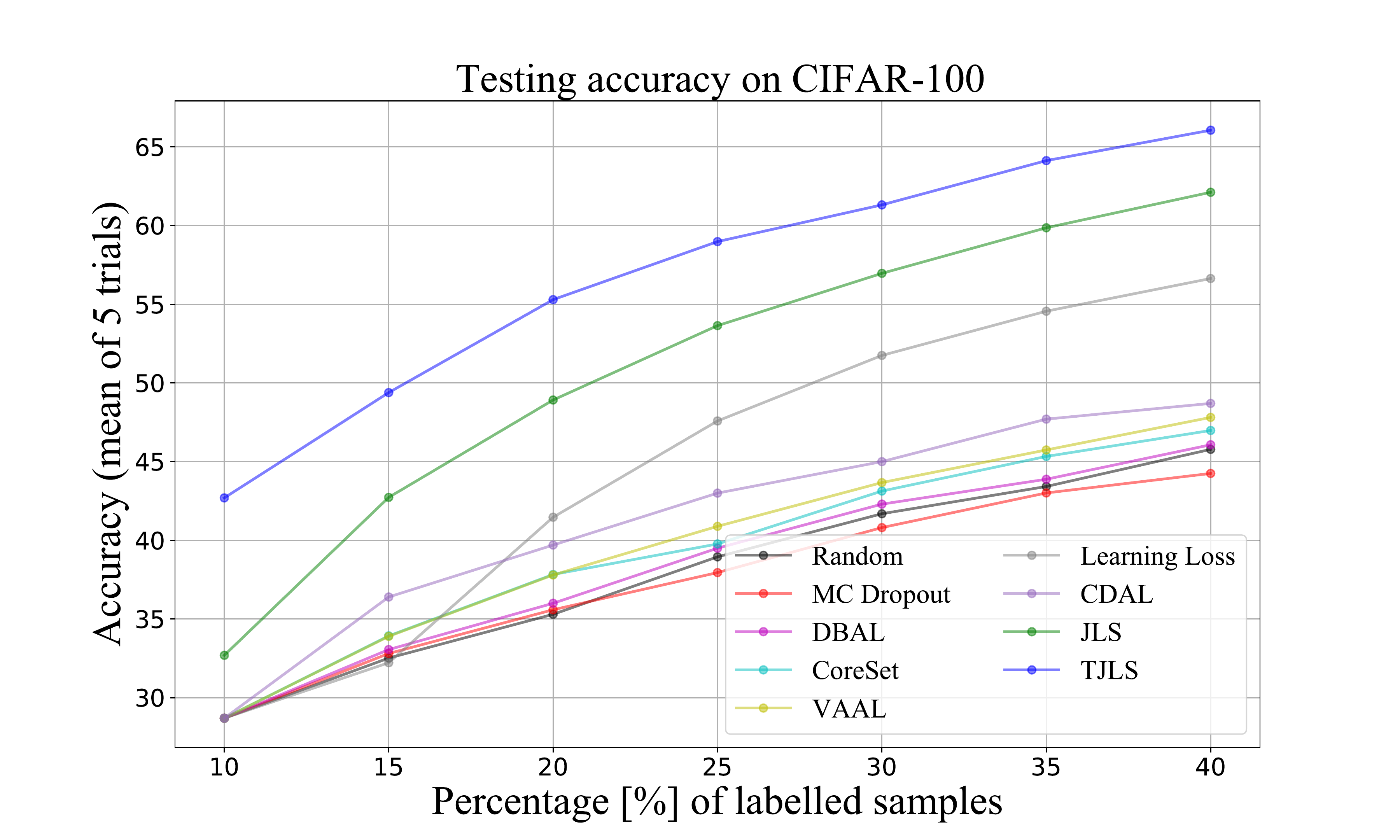}
    \includegraphics[trim=2cm 0cm 2cm 0cm, clip, width=0.32\textwidth]{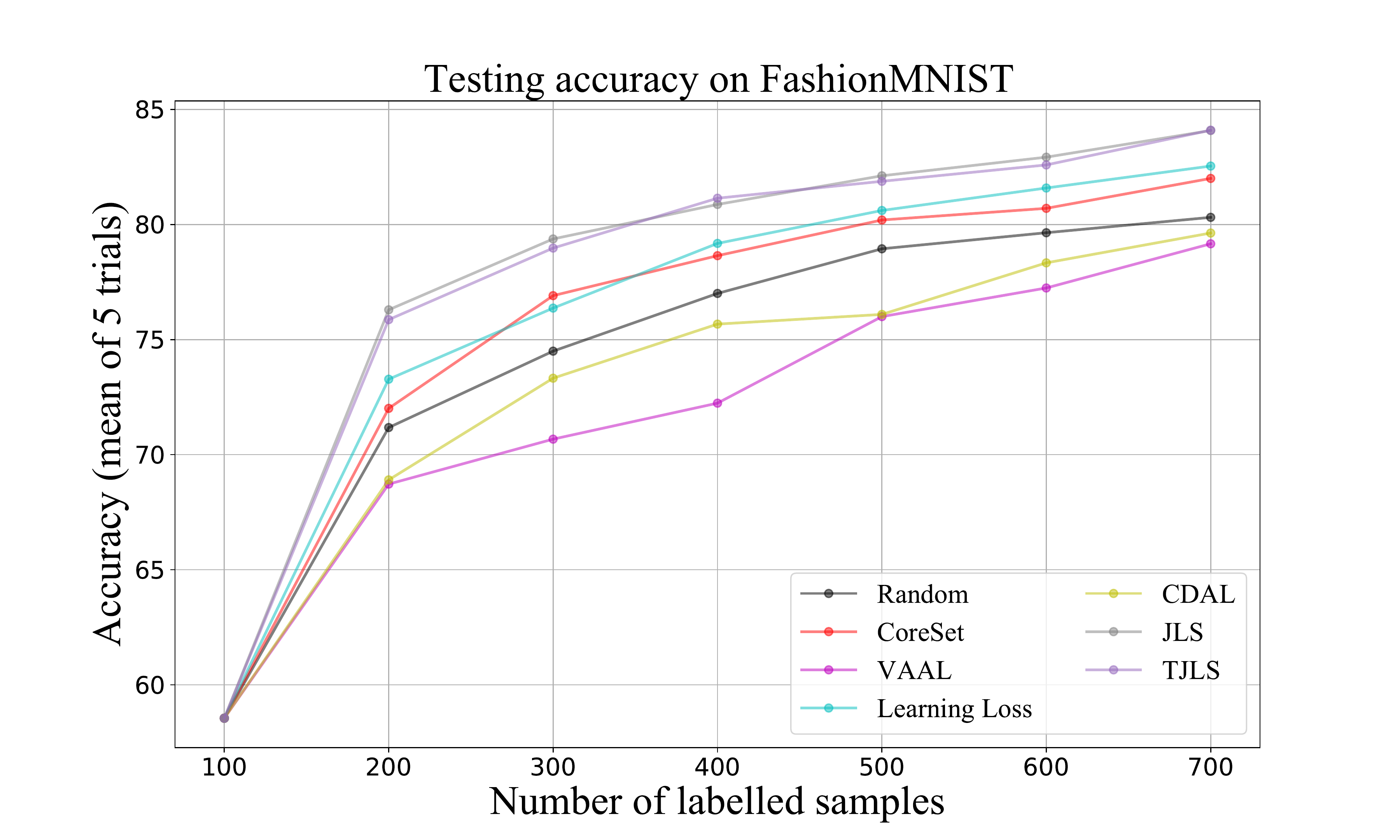}
    \caption{Quantitative evaluation on the CIFAR-10 \textbf{left}, CIFAR-100 \textbf{middle} sets with VGG-16 and  FashionMNIST (right) with ResNet-18(\textbf{right}) [Zoom in for better view]}
    \label{fig:cifar100}
\end{figure*}



The Figures from \ref{fig:cifar100} demonstrate quantitatively the top performance that our two versions (JLS and TJLS) achieve on CIFAR-10 (\textbf{left}) and CIFAR-100 (\textbf{middle}). The important thing to notice is that the proposed AL outperforms the baselines by a large margin from the very early stage. 
Also, the performance gain over the baselines is sustained even in the later stages.  This highlights the need and importance of addressing \emph{cold-start} problem
in model-based AL frameworks. Among the two variants of our model, TJLS and JLS, the former is more effective than  the latter one. This highlights the key role played by the transformer in non-local dependency modelling.  We observe a similar trend on FashionMNIST
(See Figure~\ref{fig:cifar100} (\textbf{right})) which is another popular grey-scale image classification benchmark.

\begin{table}[hbt!]
\label{tab:csal}
\centering
\begin{tabular}{l|l|l|l|l}
Comparison with SSL & 25\%           & 30\%           & 35\%           & 40\%           \\ \hline
CSAL {[}28{]}       & 67.93          & 68.97          & 69.8           & 70.51          \\
JLS                 & 70.2           & 71.3           & 72.18          & 71.56          \\
TJLS                & \textbf{70.22} & \textbf{71.62} & \textbf{72.45} & \textbf{72.14}
\end{tabular}
\caption{Comparison with the semi-supervised CSAL method on CIFAR-100 with a Wide ResNet-28 learner}
\vspace{-4mm}
\end{table}
In addition to previous figures, we compare the contemporary state-of-the-art SSL method CSAL \cite{csal}. We explicitly present in Table \ref{tab:csal} the quantitative results under the configuration of their work where a Wide ResNet-28 \cite{wideresnet} backbone is plugged. Beginning the AL selection at 25\% of CIFAR-100 data, our frameworks outperform by at least 2\% in testing accuracy over four stages without adding any augmented data. The results have been averaged from 3 trials.


\noindent \textbf{Intrinsic discussions.}
For a better understanding of our AL framework, we analyse the behaviour and visualise the latent features of the learner at the first selection stage through t-SNE\cite{Maaten08visualizingdata} distributions. Therefore, we run a fixed labelled set experiment on CIFAR-10 to represent both labelled and unlabelled after the second cycle of active learning. We deploy the ResNet-18 backbone for both JLS and TJLS. However, we also include the latent space evaluation for the learner without joint-learning. In this case, we apply CoreSet during the AL selection so that we can qualitatively compare it with our approach.

\begin{figure*}
    \centering
    \includegraphics[trim=2cm 1cm 3cm .5cm, clip, width=0.8\textwidth]{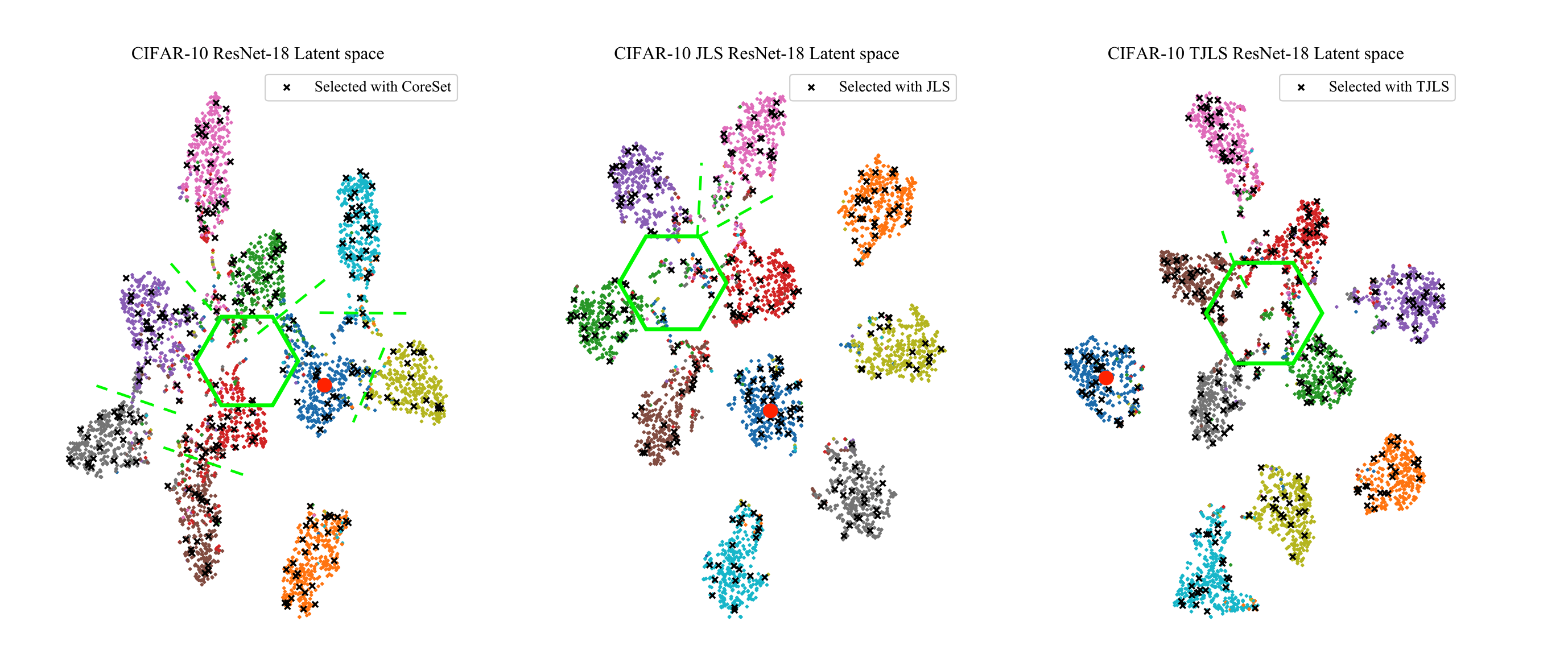}
    \caption{Intrinsic analysis of the latent space and the active learning selection [Zoom in for view]}
    \label{fig:tsne}
\end{figure*}

Figure \ref{fig:tsne} displays the three t-SNE latent spaces from the specified models: ResNet-18, JLS ResNet-18 and TJLS ResNet-18. All the images come from the available unlabelled pool. However, we already assign the 10 labels to visualise better the clusters. On this note, we also sub-sample both selected (marked with crosses) and unlabelled sets. The green hexagons added to the Figure mark the cluttered areas where some clusters are adjacent. The dotted lines also delimit presumed boundaries between classes. With these highlights, we can observe that JLS and TJLS provide more robust representations than the naive baseline. Moreover, in the cluttered areas, the sampler's uncertainty principle draws more samples than CoreSet. This will further boost the accuracy in the next training cycle.

\begin{table}[]
\centering
\small 
\label{tab:sel}
\begin{tabular}{l|c|c|c|c|c|c|c}
Selection criteria for TJLS & 10\% & 15\%           & 20\%          & 25\%          & 30\%          & 35\%          & 40\%           \\ \hline
\multicolumn{1}{c|}{TJLS + Random sampling}                                & 71.1 & 70.44          & 68.1          & 66.47         & 62.41         & 62.36         & 62.15          \\
TJLS + CoreSet {[}18{]}                                                    & -    & 72.35          & 73.55         & 74.3          & 74.5          & 74.6          & 75.3           \\
TJLS Uncertainty sampling                                                  & -    & \textbf{73.72} & \textbf{74.3} & \textbf{74.8} & \textbf{75.3} & \textbf{75.6} & \textbf{75.67}
\end{tabular}
\caption{Evaluation of different selection functions for TJLS on CIFAR-100 with ResNet-18 backbone}
\end{table}

\noindent \textbf{Ablation studies.} Although the main methodology, TJLS, relies on the visual transformer, throughout the paper, we also test the variant without it, JLS. The combined analysis leverages our motivation from the methodology. Furthermore, we re-iterate the comparison between the two and we explore the possibility of replacing the transformer with a GCN. In Table \ref{tab:abl}, we present the results of the three joint-learning architectures. Compared to Figure \ref{fig:cifar100} results, we follow the same settings and CNN backbone but we increase the number of unlabelled examples used for training by 50\%. The GCN is replacing the transformer bottleneck in the second row of Table \ref{tab:abl}. Its design is inspired from \cite{kipf2016semi} where the nodes of the graph change with input batch. Similarly to the transformer bottleneck, we want to model the higher order of representation that CNN lacks. The results in Table \ref{tab:abl} confirm the TJLS proposal by achieving the best accuracy with every labelled subset. The selection criteria of uncertainty sampling has been kept for all three variants.

\begin{table}[]
\centering
\label{tab:abl}
\small 
\begin{tabular}{c|c|c|c|c|c|c|c}
Ablation study           & 10\%           & 15\%           & 20\%           & 25\%           & 30\%           & 35\%          & 40\%          \\ \hline
JLS                      & 43             & 50             & 55.7           & 59.1           & 63             & 65            & 67.1          \\
JLS + GCN                & 45.6           & 54.23          & 59.1           & 61.9           & 64.51          & 67.4          & 69.6          \\
JLS + Transformer (TJLS) & \textbf{48.97} & \textbf{56.67} & \textbf{61.89} & \textbf{65.54} & \textbf{67.77} & \textbf{69.9} & \textbf{71.7}
\end{tabular}
\caption{ Ablation study - CIFAR-100 testing performance of the joint-learning sampling scheme (JLS), with GCN bottleneck and with TJLS (Learner VGG-16)}
\end{table}

\begin{table}[hbt!]
\label{tab:hp}
\small
\begin{tabular}{c|c|c|c|c}
\begin{tabular}[c]{@{}c@{}}TJLS \\ batch (B) \end{tabular} & 10\%  & 20\%  & 30\%  & 40\%  \\ \hline
16                                                              & 57.5  & 66.2  & 69.68 & 72.1  \\
32                                                              & 64.4  & 71.42 & 73.8  & 75.3  \\
64                                                              & 70.12 & 74.3  & 75.4  & 75.74 \\
128                                                             & 71.1  & 74.3  & 75.3  & 75.67
\end{tabular}
\hspace{0.01em}
\small 
\begin{tabular}{ccc|c|c|c|c}
\multicolumn{3}{c|}{\begin{tabular}[c]{@{}c@{}}Transformer \\ parameters\end{tabular}} & \multirow{2}{*}{10\%} & \multirow{2}{*}{20\%} & \multirow{2}{*}{30\%} & \multirow{2}{*}{40\%} \\ \cline{1-3}
depth                       & heads                       & units                      &                       &                       &                       &                       \\ \hline
1                           & 1                           & 512                        & 43.05                 & 53.9                  & 62.06                 & 65.59                 \\
1                           & 2                           & 128                        & 44.41                 & 56.05                 & 61.83                 & 66.19                 \\
2                           & 1                           & 128                        & 42.46                 & 57.04                 & 63.45                 & 66.36                
\end{tabular}

\caption{
Hyper-parameter study. (Left) ResNet-18 backbone, batch size variation. (Right) VGG-16 backbone, Transformer architectural configuration. Dataset: CIFAR-100, [mean of 3 trials], \% of labelled data }
\end{table}
\noindent \textbf{Sampler hyper-parameter study.} The sampler of our pipeline TJLS consists of two building blocks, the visual transformer and a fully-connected discriminator. We empirically evaluated both models by grid-searching the optimal architectures. However, in the discriminator case, we maintain a similar structure to the downstream task branch so that features fall under the same domain. Thus, the focus of the parameter tuning is mainly on the transformer block. Table \ref{tab:hp} presents the most meaningful results where the batch size is varied (on the left) together with depth, heads and hidden units. With the increase of the batch input size, Table \ref{tab:hp} (left), we observe that the feature's relationships are better explored. This justifies the pre-defined settings. On the right side, when changing the hidden units to 512, the gains of TJLS drop at the first selection stages. Despite this, we acknowledge that by increasing the depth or the number of heads \cite{attall}, our pipeline can achieve robust performance with different amounts of data.

\subsection{Object Detection}
Our method is generic and can be extended to other tasks simply customizing the task-specific auxiliary loss in the pipeline. Hence, we replace the categorical cross-entropy loss with SSD~\cite{ssd} loss and employed it on the object detection benchmark.
\\
\noindent \textbf{Dataset and Implementation Details.} The first work to tackle active learning for object detection is Learning Loss \cite{Yoo2019LearningLearning}. In this regard, we follow the same dataset, learner and parameter settings. Briefly, the unlabelled pool consists in 16551 images from PASCAL VOC 2007 and 2012 \cite{pascalvoc}. Compared to image classification, we start a first randomly labelled set of 1000 and we increase the budget with the same rate. However, we apply the AL selection process within 10 stages. As in \cite{Yoo2019LearningLearning, cdal}, the learner's architecture is SSD\cite{ssd} with a VGG-16\cite{Simonyan2015VeryRecognition} backbone. The visual transformer bottleneck from our joint-learning selection is positioned only on the confidence head of the SSD network.  For every AL stage evaluation, we compute the mAP metric of PASCAL VOC 2007 testing set while averaging over 5 trials.
\begin{figure*}
    \centering
    \includegraphics[trim=0cm .5cm 0cm 1cm, clip, width=0.49\textwidth]{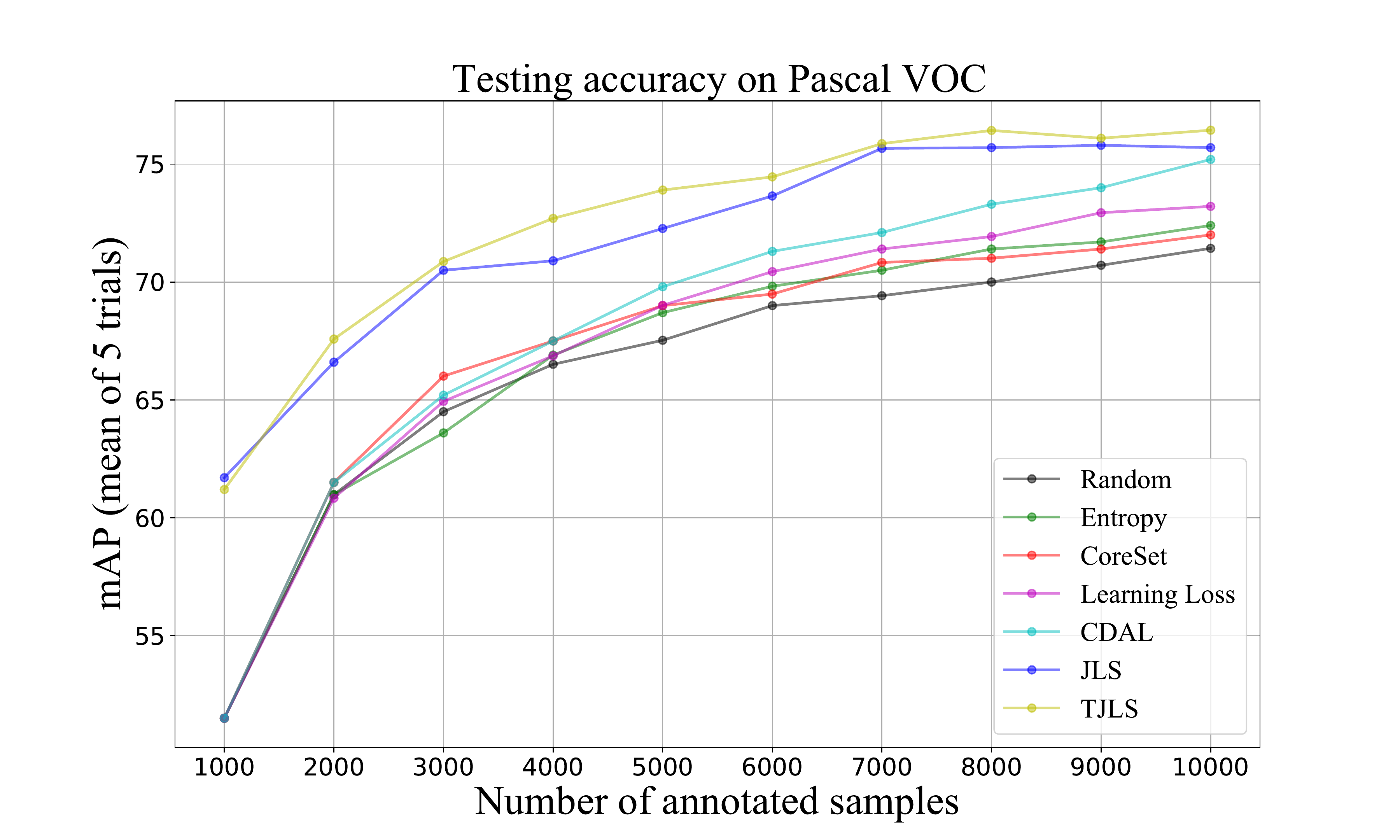}
    \includegraphics[trim=0cm .5cm 0cm 1cm, clip, width=0.49\textwidth]{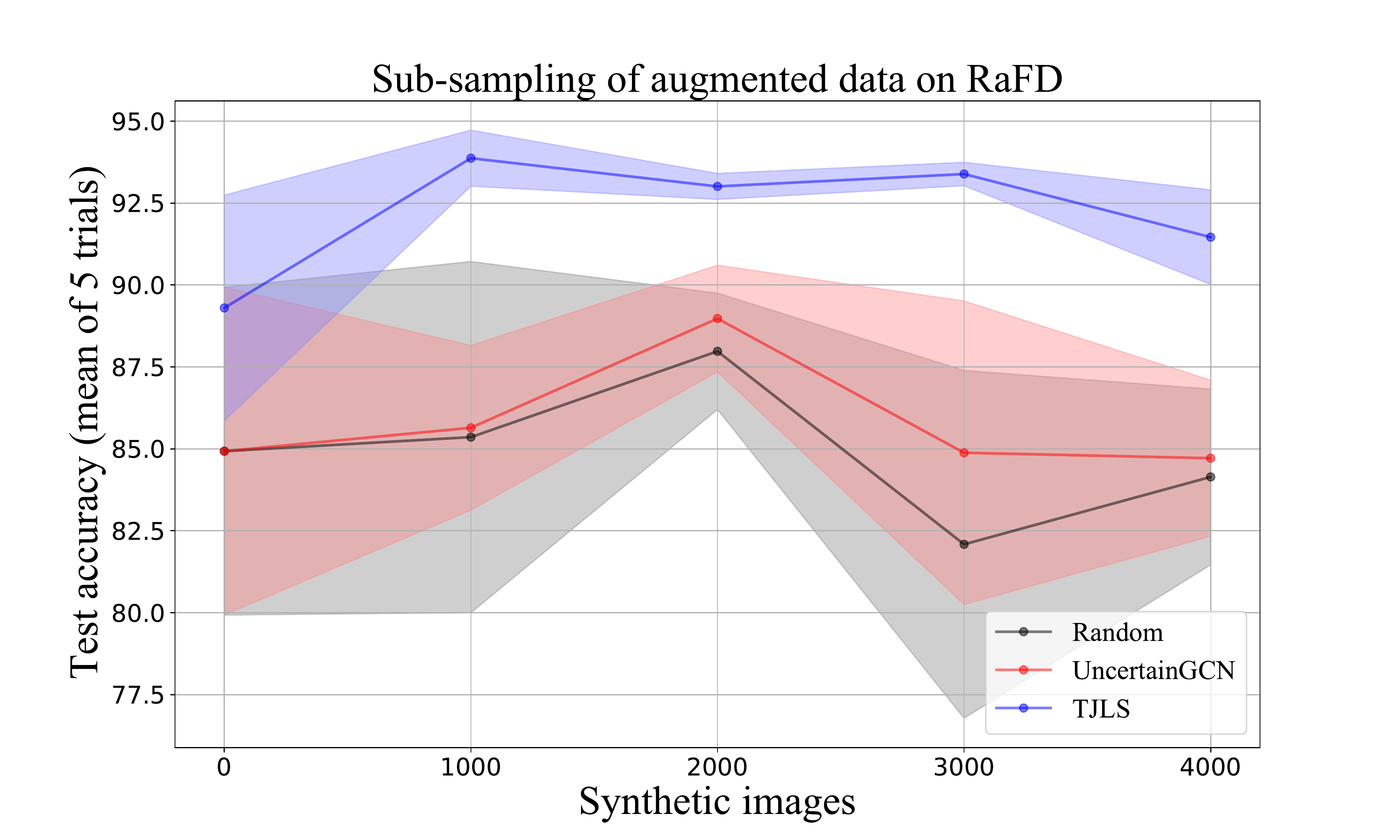}
    \caption{Quantitative evaluation on Pascal VOC 0712 dataset with SSD (\textbf{left}) and StarGAN
    synthetic data set~\textbf{(right)} [Zoom in for better view]}
    \label{fig:obj}
\end{figure*}
\\
\noindent \textbf{Quantitative analysis.} Figure \ref{fig:obj} \textbf{(left)} illustrates the comparisons of the proposed baselines on the object detection experiment. The trends of JLS and TJLS behave similarly to the image classification point. They show a great level of generalisation from the cold-start at 62.5\% mAP while having a 10\% performance gain over the other baselines. The uncertainty-based selection is appropriate with the learner's representation through all the 10 selection stages. Therefore, both JLS and TJLS saturate with top performance from the 7th cycle at over 76\% mAP. Our method outperforms the previous state-of-the-art Learning Loss and CDAL.
However, between the two proposed variants, TJLS provides non-local interactions within the batch. This advantage is reflected quantitatively against JLS in this task as well.

\subsection{Subsampling synthetic data} Sub-sampling synthetic data to augment the real data is an active research~\cite{ugcn,bhattarai2020sampling}. 
Following the experimental setup of Caramalau \etal~\cite{ugcn} for sub-setting synthetic data,  we employed our pipeline to select the face expression 
synthetic data generated by StarGAN~\cite{choi2018stargan}. The selected synthetic data were augmented with the RaFD\footnote{http://www.socsci.ru.nl:8180/RaFD2/RaFD} real training data to train to a model
for face expression classification task. Figure~\ref{fig:obj} (\textbf{right}) shows the performance comparison. 
Our method surpasses the performance of existing arts by a large margin. 


\section{Conclusions and Limitations}
In this paper, we present a novel model-based active learning. Our contributions are the adaptation of Visual Transformer to address non-local dependencies between all the examples and exploitation of the unlabelled data by jointly minimizing the task-aware objective. Our extensive empirical and qualitative analysis on
multiple benchmarks demonstrate the efficacy of the proposed method compared to the existing method.
The main fallback to address in the proposed method is scalability. We acknowledge that introducing the visual transformer in TJLS increases the number of target model parameters. Despite this, our proposal relies on the current and future technological advancements where there has already been continuous growth. Another caveat consists of the restriction of the batch self-attention block. Some architectures might require smaller batch sizes where the benefits of TJLS can be affected. From recent works \cite{vt,vi}, it has been shown that visual transformers demand a big corpus of data. However, by including the unlabelled examples as part of the TJLS training, we satisfy this requirement.  In future work, we would like to explore our method to efficiently handle high-resolution data. 



\noindent \textbf{ Broader impact:}
 Active learning is a dynamic and important research topic. Our contribution lies in the methodology
of active learning. We believe that the research proposed in this paper can be further applied where
large-scale data and annotation presents an issue. Thus, this might serve to fields like medical imaging,
robotics and many other. Moreover, integrating TJLS as a sampling framework would
yield greater performance in a limited labelled data scenario. Our method opens a new direction in
the active learning research being sustained by state-of-the-art results.

\clearpage

\bibliographystyle{ieee_fullname}
\bibliography{neurips2021}

\begin{thebibliography}{10}\itemsep=-1pt

\bibitem{cdal}
Sharat Agarwal, Himanshu Arora, Saket Anand, and Chetan Arora.
\newblock Contextual diversity for active learning.
\newblock In {\em ECCV}, 2020.

\bibitem{BeluchBcai2018TheClassification}
William~H Beluch~Bcai, Andreas N{\"{u}}rnberger, and Jan M~Köhler Bcai.
\newblock {The power of ensembles for active learning in image classification}.
\newblock In {\em CVPR}, 2018.

\bibitem{bhattarai2020sampling}
Binod Bhattarai, Seungryul Baek, Rumeysa Bodur, and Tae-Kyun Kim.
\newblock Sampling strategies for gan synthetic data.
\newblock In {\em ICASSP}, 2020.

\bibitem{caramalau2021active}
Razvan Caramalau, Binod Bhattarai, and Tae-Kyun Kim.
\newblock Active learning for bayesian 3d hand pose estimation.
\newblock In {\em WACV}, 2021.

\bibitem{ugcn}
Razvan Caramalau, Binod Bhattarai, and Tae-Kyun Kim.
\newblock Sequential graph convolutional network for active learning.
\newblock In {\em CVPR}, 2021.

\bibitem{objdet2020tf}
Nicolas Carion, Francisco Massa, Gabriel Synnaeve, Nicolas Usunier, Alexander
  Kirillov, and Sergey Zagoruyko.
\newblock End-to-end object detection with transformers.
\newblock In {\em ECCV}, 2020.

\bibitem{caruana1997multitask}
Rich Caruana.
\newblock Multitask learning.
\newblock {\em Machine learning}, 28(1):41--75, 1997.

\bibitem{choi2018stargan}
Yunjey Choi, Minje Choi, Munyoung Kim, Jung-Woo Ha, Sunghun Kim, and Jaegul
  Choo.
\newblock Stargan: Unified generative adversarial networks for multi-domain
  image-to-image translation.
\newblock In {\em CVPR}, 2018.

\bibitem{bert}
Jacob Devlin, Ming-Wei Chang, Kenton Lee, and Kristina Toutanova.
\newblock {BERT}: Pre-training of deep bidirectional transformers for language
  understanding.
\newblock In {\em Proceedings of the 2019 Conference of the North {A}merican
  Chapter of the Association for Computational Linguistics: Human Language
  Technologies, Volume 1 (Long and Short Papers)}, June 2019.

\bibitem{vi}
Alexey Dosovitskiy, Lucas Beyer, Alexander Kolesnikov, Dirk Weissenborn,
  Xiaohua Zhai, Thomas Unterthiner, Mostafa Dehghani, Matthias Minderer, Georg
  Heigold, Sylvain Gelly, Jakob Uszkoreit, and Neil Houlsby.
\newblock An image is worth 16x16 words: Transformers for image recognition at
  scale, 2020.

\bibitem{sslalspeech}
Thomas Drugman, Janne Pylkk{\"o}nen, and Reinhard Kneser.
\newblock Active and semi-supervised learning in asr: Benefits on the acoustic
  and language models.
\newblock In {\em INTERSPEECH}, 2016.

\bibitem{pascalvoc}
Mark Everingham, Luc Gool, Christopher~K. Williams, John Winn, and Andrew
  Zisserman.
\newblock The pascal visual object classes (voc) challenge.
\newblock {\em International Journal of Computer Vision}, 2010.

\bibitem{Gal2016DropoutGhahramani}
Yarin Gal and Zoubin Ghahramani.
\newblock {Dropout as a Bayesian Approximation: Representing Model Uncertainty
  in Deep Learning}.
\newblock In {\em ICML}, 2016.

\bibitem{Gal2017DeepDatab}
Yarin Gal, Riashat Islam, and Zoubin Ghahramani.
\newblock {Deep Bayesian Active Learning with Image Data}.
\newblock In {\em ICML}, 2017.

\bibitem{csal}
Mingfei Gao, Zizhao Zhang, Guo Yu, Sercan Arık, Larry Davis, and Tomas
  Pfister.
\newblock Consistency-based semi-supervised active learning: Towards minimizing
  labeling cost.
\newblock In {\em ECCV}, pages 510--526, 2020.

\bibitem{objdet2020sota}
Golnaz {Ghiasi}, Yin {Cui}, Aravind {Srinivas}, Rui {Qian}, Tsung-Yi {Lin},
  Ekin~D. {Cubuk}, Quoc~V. {Le}, and Barret {Zoph}.
\newblock {Simple Copy-Paste is a Strong Data Augmentation Method for Instance
  Segmentation}.
\newblock {\em arXiv e-prints}, 2020.

\bibitem{inductivebias}
Anirudh {Goyal} and Yoshua {Bengio}.
\newblock {Inductive Biases for Deep Learning of Higher-Level Cognition}.
\newblock {\em arXiv e-prints}, 2020.

\bibitem{Har-Peled2007}
Sariel Har-Peled and Akash Kushal.
\newblock Smaller coresets for k-median and k-means clustering.
\newblock In {\em SCG}, 2005.

\bibitem{he2016deep}
Kaiming He, Xiangyu Zhang, Shaoqing Ren, and Jian Sun.
\newblock Deep residual learning for image recognition.
\newblock In {\em CVPR}, 2016.

\bibitem{lstm}
Sepp Hochreiter and J\"{u}rgen Schmidhuber.
\newblock Long short-term memory.
\newblock {\em Neural Comput.}, 1997.

\bibitem{Houlsby2011BayesianLearning}
Neil Houlsby, Ferenc Husz{\'{a}}r, Zoubin Ghahramani, and Máté Lengyel.
\newblock {Bayesian Active Learning for Classification and Preference
  Learning}, 2011.
\newblock 1112.5745v1.

\bibitem{kipf2016semi}
Thomas~N Kipf and Max Welling.
\newblock Semi-supervised classification with graph convolutional networks.
\newblock In {\em ICLR}, 2017.

\bibitem{Kirsch2019BatchBALD:Learning}
Andreas Kirsch, Joost Van~Amersfoort, and Yarin Gal.
\newblock {BatchBALD: Efficient and Diverse Batch Acquisition for Deep Bayesian
  Active Learning}.
\newblock In {\em NeurIPS}, 2019.

\bibitem{near2}
Aryeh Kontorovich, Sivan Sabato, and Ruth Urner.
\newblock Active nearest-neighbor learning in metric spaces.
\newblock In {\em NeurIPS}, 2016.

\bibitem{cifar}
Alex Krizhevsky.
\newblock Learning multiple layers of features from tiny images.
\newblock {\em University of Toronto}, 05 2012.

\bibitem{krizhevsky2012imagenet}
Alex Krizhevsky, Ilya Sutskever, and Geoffrey~E Hinton.
\newblock Imagenet classification with deep convolutional neural networks.
\newblock In {\em NeurIPS}, 2012.

\bibitem{curmat}
Changsheng Li, Xiangfeng Wang, Weishan Dong, Junchi Yan, Qingshan Liu, and
  Hongyuan Zha.
\newblock Joint active learning with feature selection via cur matrix
  decomposition.
\newblock {\em IEEE Transactions on Pattern Analysis and Machine Intelligence},
  2019.

\bibitem{ssd}
Wei Liu, Dragomir Anguelov, Dumitru Erhan, Christian Szegedy, Scott Reed,
  Cheng-Yang Fu, and Alexander Berg.
\newblock Ssd: Single shot multibox detector.
\newblock In {\em ECCV}, 2016.

\bibitem{Pinsler2019BayesianApproximation}
Robert Pinsler, Jonathan Gordon, Eric Nalisnick, and José Miguel
  Hernandez-Lobato.
\newblock {Bayesian Batch Active Learning as Sparse Subset Approximation}.
\newblock In {\em NeurIPS}, 2019.

\bibitem{pu2016variational}
Yunchen Pu, Zhe Gan, Ricardo Henao, Xin Yuan, Chunyuan Li, Andrew Stevens, and
  Lawrence Carin.
\newblock Variational autoencoder for deep learning of images, labels and
  captions.
\newblock In {\em NeurIPS}, 2016.

\bibitem{Sener2017ActiveApproach}
Ozan Sener and Silvio Savarese.
\newblock {Active Learning for Convolutional Neural Networks: A Core-set
  approach}.
\newblock In {\em ICLR}, 2018.

\bibitem{settles.tr09}
Burr Settles.
\newblock Active learning literature survey.
\newblock Computer Sciences Technical Report 1648, University of
  Wisconsin--Madison, 2009.

\bibitem{entropy}
C.~E. Shannon.
\newblock A mathematical theory of communication.
\newblock {\em The Bell System Technical Journal}, 1948.

\bibitem{Simonyan2015VeryRecognition}
Karen Simonyan and Andrew Zisserman.
\newblock {Very Deep Convolutional Network for Large-scale image recognition}.
\newblock In {\em ICLR}, 2015.

\bibitem{Sinha2019VariationalLearning}
Samarth Sinha, Sayna Ebrahimi, and Trevor Darrell.
\newblock {Variational Adversarial Active Learning}.
\newblock In {\em ICCV}, 2019.

\bibitem{coresvm}
Ivor~W. Tsang, James~T. Kwok, and Pak-Ming Cheung.
\newblock Core vector machines: Fast svm training on very large data sets.
\newblock {\em JMLR.}, 2005.

\bibitem{Maaten08visualizingdata}
Laurens van~der Maaten and Geoffrey Hinton.
\newblock Visualizing data using t-sne, 2008.
\newblock JMLR.

\bibitem{attall}
Ashish Vaswani, Noam Shazeer, Niki Parmar, Jakob Uszkoreit, Llion Jones,
  Aidan~N Gomez, \L~ukasz Kaiser, and Illia Polosukhin.
\newblock Attention is all you need.
\newblock In {\em NeurIPS}, 2017.

\bibitem{wolf}
Gert Wolf.
\newblock Facility location: concepts, models, algorithms and case studies.
\newblock In {\em Contributions to Management Science}, 2011.

\bibitem{vt}
Bichen Wu, Chenfeng Xu, Xiaoliang Dai, Alvin Wan, Peizhao Zhang, Zhicheng Yan,
  Masayoshi Tomizuka, Joseph Gonzalez, Kurt Keutzer, and Peter Vajda.
\newblock Visual transformers: Token-based image representation and processing
  for computer vision, 2020.

\bibitem{Xiao2017Fashion-MNIST:Algorithms}
Han Xiao, Kashif Rasul, and Roland Vollgraf.
\newblock {Fashion-MNIST: a Novel Image Dataset for Benchmarking Machine
  Learning Algorithms}, 2017.
\newblock 1708.07747v2.

\bibitem{Yoo2019LearningLearning}
Donggeun Yoo and In~So Kweon.
\newblock {Learning Loss for Active Learning}.
\newblock In {\em CVPR}, 2019.

\bibitem{wideresnet}
Sergey Zagoruyko and Nikos Komodakis.
\newblock Wide residual networks.
\newblock In {\em BMVC}, 2016.

\bibitem{Zhang2020State-RelabelingLearning}
Beichen Zhang, Liang Li, Shijie Yang, Shuhui Wang, Zheng-Jun Zha, and Qingming
  Huang.
\newblock {State-Relabeling Adversarial Active Learning}.
\newblock In {\em CVPR}, 2020.

\bibitem{objdet2018sota}
Shifeng Zhang, Longyin Wen, Xiao Bian, Zhen Lei, and Stan Li.
\newblock Single-shot refinement neural network for object detection.
\newblock In {\em CVPR}, 2018.

\end{thebibliography}

\clearpage
\appendix
\section{Supplementary Material}
\setcounter{table}{0}
\counterwithin{table}{section}
\subsection{Standard deviations in the quantitative evaluations}
For a better clarity, in our figures of image classification (Figure \ref{fig:cifar100}) and object detection (Figure \ref{fig:obj} left) we excluded the standard deviation representation. We kept only the mean value to avoid overlapping. However, we present these values in Table \ref{tab:app}.
\begin{table}[hbt!]
\label{tab:app}
\centering
\begin{tabular}{c|c|c|c|c|c|c|c}
Selection cycle         & 1   & 2   & 3   & 4   & 5   & 6   & 7   \\ \hline
{[}CIFAR-10{]} JLS      & .21 & .09 & .3  & .11 & .2  & .19 & .33 \\
{[}CIFAR-10{]} TJLS     & .16 & .2  & .23 & .12 & .1  & .09 & .15 \\
{[}CIFAR-100{]} JLS     & .04 & .01 & .22 & .18 & .22 & .19 & .18 \\
{[}CIFAR-100{]} TJLS    & .1  & .16 & .33 & .17 & .31 & .15 & .31 \\
{[}FashionMNIST{]} JLS  & .44 & .64 & .55 & .13 & .47 & .25 & .75 \\
{[}FashionMNIST{]} TJLS & .95 & .75 & .46 & .4  & .19 & .58 & .7  \\
{[}PASCAL VOC{]} JLS    & .06 & .04 & .03 & .06 & .02 & .01 & .05 \\
{[}PASCAL VOC{]} TJLS   & .1  & .02 & .04 & .08 & .04 & .02 & .07
\end{tabular}
\caption{Standard deviation of the JLS/TJLS qualitative results on CIFAR-10/100, FashionMNIST and Pascal VOC}
\end{table}

We can observe that the deviations in most experiments are relatively low. This robustness happens due to the high degree of generalisation while training with our proposed pipeline. The measurements are in decimals of the 
testing accuracy/mAP percentages.

\subsection{Experiments compute resources}

We conduct all our experiments in Python3 with the PyTorch deep learning library. To speed up the running process, we train the models on Graphical Processing Units (GPUs). For image classification, we can fit any of the presented architecture on a single NVIDIA 1080Ti GPU with 11GB memory. However, the object detection models are larger and we parallelised the processes on two GPUs. More details regarding the increase of parameters in our JLS and TJLS frameworks are enlisted in Table \ref{tab:param}. 

\begin{table}[hbt!]
\centering
\label{tab:param}
\begin{tabular}{c|c|c|c}
\begin{tabular}[c]{@{}c@{}}Model / Number \\ of parameters\end{tabular} & Baseline   & JLS        & TJLS       \\ \hline
VGG-16                                                                  & 14,765,988 & 15,029,157 & 15,620,005 \\
ResNet-18                                                               & 11,220,132 & 11,483,301 & 12,074,149 \\
SSD                                                                     & 26,285,486 & 26,468,859 & 29,982,859
\end{tabular}
\caption{Number of parameters of JLS and TJLS samplers added to the VGG-16, ResNet-18, SSD backbones}
\end{table}
\clearpage

\end{document}